\documentclass[letterpaper]{article} 
\usepackage{aaai2027}  
\usepackage[hyphens]{url}  
\usepackage{graphicx} 
\usepackage{natbib}  
\usepackage{caption} 
\usepackage{algorithm}
\usepackage{algorithmic}
\usepackage{amsmath}
\usepackage{amssymb} 
\usepackage{booktabs}
\usepackage{fvextra}

\DefineVerbatimEnvironment{PromptBlock}{Verbatim}{
  fontsize=\scriptsize,
  breaklines=true,
  breakanywhere=true,
  breaksymbolleft={}
}

\usepackage{xcolor}
\usepackage{colortbl}
\usepackage{tabularx}
\definecolor{ourshade}{gray}{0.93}
\title{CausalOPD: First-Wrong-Step Supervision\\for Distilling Causal Chain Reasoning}
\author{Jian Zhang, Bingyi Wang, Yizhi Liu}
\affiliations{Zhejiang University}
\nocopyright
\begin{document}

\maketitle

\begin{abstract}
Many critical reasoning tasks, including clinical diagnosis, legal judgment reasoning, and industrial fault diagnosis, share a common structure: a causal attribution is reached through a step-dependent chain in which each stage constrains the next. An early error propagates through the chain, and a correct conclusion reached via invalid reasoning evades outcome-only evaluation. Large language models excel at such multi-step causal reasoning, yet privacy, latency, and controllability requirements motivate distilling them into locally deployable small models. Standard trajectory-imitation methods do not correct process-level errors on the student's own rollout distribution, allowing early-stage mistakes to compound through the causal chain. We propose CausalOPD, a curriculum online process distillation framework for step-dependent causal reasoning. CausalOPD first cold-starts the student with reasoning trajectories from a knowledge-augmented teacher grounded in domain-specific causal rules, entity relationships, and structural constraints. The student then generates on-policy trajectories on training cases; the teacher localizes the first wrong step, the earliest transition that verifiably contradicts the available knowledge constraints. From the verified prefix and localized error, short-horizon reinforcement learning targets the specific failure rather than re-optimizing the entire response. A causal-stage curriculum progressively advances the training focus from evidence-level through mechanism-level to conclusion-level error repair, following the order in which errors propagate through the causal chain. Across three domains, CausalOPD improves average path correctness by 23.4 percentage points over sequence-level OPD and reduces the right-label--wrong-reasoning (RLWR) rate from 15.7\% to 4.4\%. The domain-specific 8B students surpass both proprietary references on path correctness in all three domains.
\end{abstract}

\section{Introduction}

Large language models (LLMs) are increasingly applied to multi-step causal reasoning, where conclusions are derived through chains of interdependent inferences. Such reasoning underlies decision-critical applications, including clinical diagnosis, legal judgment reasoning, and industrial fault diagnosis \citep{singhal2023large,guha2023legalbench,appliedenergy2024fault,aei2026fault}. In these settings, the reasoning path itself carries decision value: practitioners must assess whether relevant evidence has been identified, whether the inferred mechanism is supported, and whether the final attribution follows from that mechanism. A correct conclusion reached through an invalid path is therefore insufficient. Meanwhile, requirements for data privacy, inference latency, and controllability often limit the use of proprietary LLMs served through external APIs. This creates a practical need to transfer their causal reasoning capabilities to smaller, locally deployable models.

Knowledge distillation provides a natural framework for this transfer \citep{hinton2015distilling}. Existing approaches have improved reasoning supervision by aligning it more closely with student behavior and by increasing its granularity. Conventional methods train students to imitate teacher-generated reasoning trajectories \citep{hsieh2023distilling,muhebwa2025causal}. Sampled from the teacher, such trajectories do not expose the teacher to errors arising from the student's inference distribution. On-policy distillation (OPD) addresses this mismatch by supervising trajectories generated by the student itself \citep{gu2024minillm,agarwal2024onpolicy}. Standard OPD objectives primarily align the student's output distribution with the teacher's distribution on these trajectories, but do not explicitly identify which transition first renders a causal path invalid.

In parallel, process supervision has refined the granularity of reasoning feedback. Outcome rewards assign a single score to a complete trajectory by checking its final conclusion \citep{guo2025deepseekr1}. A multi-step trajectory may nevertheless reach a correct conclusion despite containing invalid intermediate inferences. Trajectory-level feedback can neither reliably distinguish such a trajectory from one supported by a valid reasoning path nor attribute failure to the transition that caused it. This creates a credit-assignment gap: optimization may reward trajectories with correct conclusions even when their intermediate causal transitions are invalid. Process reward models (PRMs) mitigate this problem by assessing intermediate steps using human annotations \citep{uesato2022solving,lightman2024lets} or continuation-based estimates of eventual success \citep{wang2024mathshepherd}.

These developments make supervision more distribution-aligned and fine-grained, but neither property alone captures the dependency structure of causal reasoning. Existing approaches treat reasoning steps as locally scorable units; in a causal chain, what follows the first error is largely its consequence: evidence identification constrains mechanism inference, which in turn constrains conclusion attribution. This dependency creates challenges concerning the reliability, localization, and progression of supervision.

\begin{itemize}
    \item The reliability challenge arises from the source of process judgments. Even strong LLMs can fail to apply formal causal rules consistently \citep{jin2023cladder}, while supervision inherited solely from teacher outputs may reproduce unsupported or erroneous reasoning in the student \citep{gudibande2024imitation}. Process judgments should therefore be grounded in explicit domain knowledge rather than solely in the teacher's implicit parameters.

    \item The localization challenge follows from error propagation. An early error can invalidate subsequent steps even when they remain internally coherent under a false premise. Independently scoring steps or re-optimizing the complete response may therefore conflate an originating error with its downstream consequences. Effective correction must identify the earliest transition that can be shown to violate the available constraints.

    \item The progression challenge concerns the order in which reasoning competence is developed. Reliable mechanism inference presupposes correct evidence identification, and reliable conclusion attribution presupposes a valid mechanism. Applying supervision without respecting these prerequisites overlooks the direction in which errors propagate through the chain.
\end{itemize}

Together, these challenges motivate a dependency-aware principle for process correction: optimization should begin at the \emph{first wrong step}, operationally defined as the earliest transition that verifiably violates the available domain knowledge and causal constraints. This definition is coverage-aware: the absence of a detected violation does not imply that a step is universally correct. Localizing this transition preserves the verified prefix, separates a detected originating error from its propagated consequences, and reduces the optimization horizon from the complete response to a targeted reasoning segment. The same dependency structure suggests that correction should progress from evidence identification to mechanism inference and conclusion attribution. Without such supervision, a student may reach a correct conclusion through an invalid path, creating an outcome--process inconsistency that outcome-only evaluation cannot detect and that may indicate shortcut behavior \citep{geirhos2020shortcut,turpin2023language}.

We propose \textbf{CausalOPD}, a curriculum online process distillation framework that applies knowledge-grounded process supervision to student-generated causal reasoning trajectories. CausalOPD makes the following contributions:
\begin{itemize}
    \item We introduce knowledge-grounded first-wrong-step supervision: a knowledge-augmented teacher verifies student-generated trajectories against domain-specific causal rules, entity relationships, and structural constraints, and localizes the earliest transition that verifiably violates them, separating the originating error from its downstream consequences.

    \item We develop localized online process optimization that retains the verified prefix, optimizes only the affected suffix with short-horizon reinforcement learning, and advances correction from evidence identification through mechanism inference to conclusion attribution according to their causal prerequisites.

    \item Across industrial, clinical, and legal benchmarks, CausalOPD consistently improves reasoning-path correctness and conclusion accuracy over trajectory SFT and a sequence-level on-policy distillation baseline. The domain-specific 8B students additionally surpass both proprietary references on path correctness in all three domains and their zero-shot teacher on every metric.
\end{itemize}

\section{Related Work}

\subsection{On-Policy Reasoning Distillation}

Knowledge distillation was originally developed to transfer the predictive behavior of a large teacher to a smaller student \citep{hinton2015distilling}. With the emergence of chain-of-thought reasoning, this paradigm expanded from matching output distributions to imitating teacher-generated rationales, allowing students to learn intermediate reasoning alongside final answers \citep{hsieh2023distilling,muhebwa2025causal}. These trajectories, however, are sampled from the teacher and may not reflect the errors made by the student at inference time. On-policy distillation addresses this mismatch by applying teacher supervision to student-generated trajectories, typically by aligning the student's output distribution with that of the teacher \citep{gu2024minillm,agarwal2024onpolicy}. Yet errors in student-generated trajectories still raise challenges of reliable evaluation, structural localization, and progressive correction.

\begin{figure*}[t]
\centering
\includegraphics[width=\textwidth]{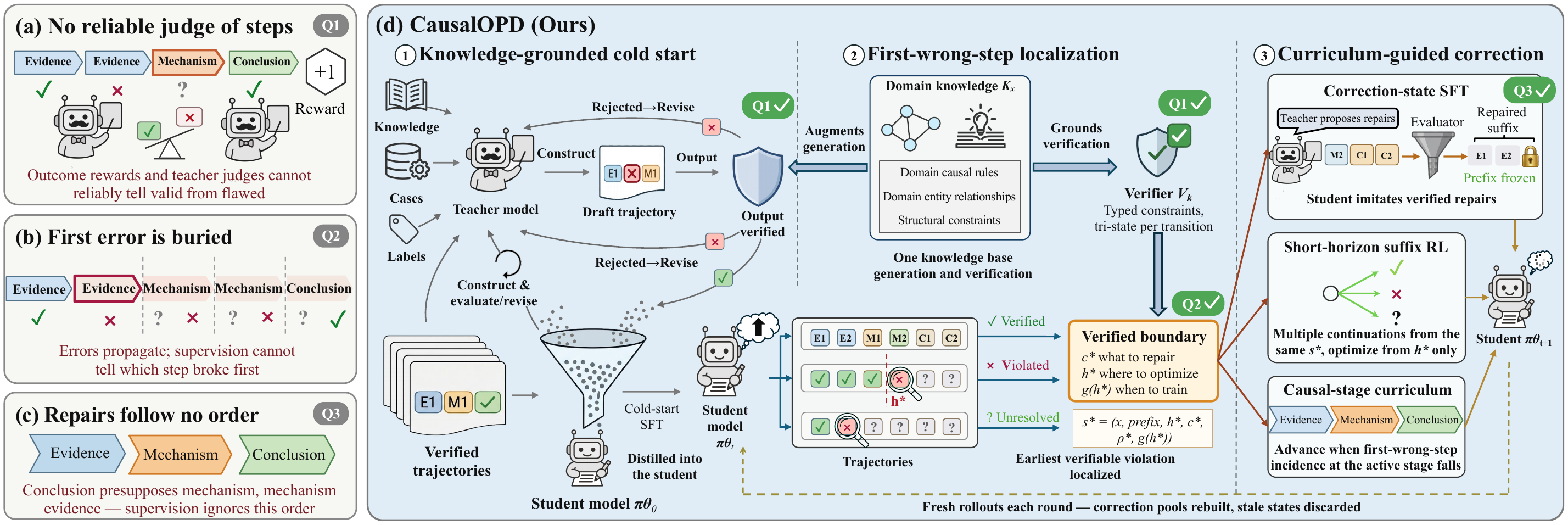}
\caption{(a--c) Three challenges in causal-chain distillation; (d) the CausalOPD training pipeline.}
\label{fig:overview}
\end{figure*}

\subsection{Knowledge-Enhanced Distillation}

Teacher-generated rationales provide rich supervision, but their reliability is bounded by the teacher's knowledge and reasoning ability. External knowledge has therefore been incorporated into distillation. KARD, for example, fine-tunes small models to generate teacher rationales conditioned on passages retrieved from an external knowledge base, offloading knowledge that a small model cannot reliably memorize \citep{kang2023knowledge}. Related work has also distilled structured causal explanations from strong teachers into smaller models \citep{muhebwa2025causal}. These methods establish the value of external knowledge in distillation, yet such knowledge primarily supports generation rather than judgment, and stronger generation does not by itself ensure reliable evaluation of student-generated reasoning. Even strong models can fail to apply formal causal rules consistently \citep{jin2023cladder}, and imitating a teacher's style does not ensure comparable factuality \citep{gudibande2024imitation}. Logic-LM uses symbolic solvers to check translated programs and derive final answers \citep{pan2023logiclm}, but does not verify individual reasoning transitions within student-generated trajectories. This leaves the \emph{reliability challenge} open: how can explicit domain knowledge support not only generation but also verification, judging student-generated reasoning transitions against causal rules, entity relationships, and structural constraints?

\subsection{Process Supervision and Correction}

Outcome-level supervision evaluates a reasoning trajectory only through its final conclusion, which may be reached through invalid intermediate inferences and thus reveals little about where the process fails \citep{uesato2022solving,guo2025deepseekr1}. Process reward models address this credit-assignment problem by scoring intermediate steps with human annotations \citep{uesato2022solving,lightman2024lets} or continuation-based estimates of eventual success \citep{wang2024mathshepherd}, yet reliable process-error detection remains difficult even for strong models \citep{zheng2024processbench,zhang2025lessons}. Moving from evaluation to revision, self-correction methods refine previous responses but are often inconsistent without reliable external feedback \citep{madaan2023selfrefine,huang2024large}, and SCoRe trains correction behavior across multiple attempts via reinforcement learning \citep{kumar2025training}. Closer approaches localize optimization: GLoRe conditions refinement on a first-error location predicted by a learned reward model \citep{havrilla2024glore}, whereas R$^3$ starts reinforcement learning from states along correct demonstrations \citep{xi2024r3}. Neither establishes the correction boundary by verifying the student's own reasoning transitions against explicit domain constraints. This leaves the \emph{localization challenge} open: how can supervision place this boundary at the earliest verifiably violated transition, and how can correction preserve the verified prefix while targeting only the affected suffix?

\subsection{Competence-Aware Curricula}

Curriculum learning organizes training from easier to more difficult examples \citep{bengio2009curriculum}. Later approaches construct dynamic curricula using sample difficulty, data quality, noise, or model competence \citep{soviany2022curriculum}. Such curricula make training more responsive to student competence, but primarily determine which examples should be presented at each stage; they do not explicitly determine which component of a dependent reasoning process should be repaired first. This leaves the \emph{progression challenge} open: how should optimization advance when mechanism inference presupposes reliable evidence identification and conclusion attribution presupposes a valid mechanism?

\paragraph{Summary.}
Prior work separately improves rollout alignment, knowledge grounding, feedback granularity, and training scheduling, but their joint treatment in causal-chain distillation remains underexplored. \textbf{CausalOPD} couples them through one verified boundary: the violated constraint specifies what to repair, its position fixes where optimization begins, and its causal stage schedules when the correction is trained.

\section{Method}

Figure~\ref{fig:overview} summarizes CausalOPD, which supervises student-generated trajectories at the \emph{first wrong step}. A single knowledge-derived localization result coordinates the framework: the violated constraint $c^{*}$ determines what to repair, its position $h^{*}$ determines where optimization begins, and its stage $g(h^{*})$ determines when the correction state is prioritized. Training combines knowledge-grounded cold start, correction-state fine-tuning, and curriculum-guided short-horizon reinforcement learning with per-round fresh rollouts.

\subsection{Problem Formulation}

We first formalize the asymmetric training setting and the localization interface that all subsequent training phases consume. Throughout, causal reasoning denotes step-dependent causal attribution under documented mechanisms, not interventional or counterfactual inference. Each case provides observations $x$ and inference-time system/context metadata $\mu_x$ to every method; in industrial diagnosis, these cover sensor observations, system topology, control context, and observability. The teacher additionally accesses the training-only privileged context $p_x=(y,\mathcal{K}_x)$: the reference conclusion $y$ and general mechanism knowledge $\mathcal{K}_x$ (applicable causal rules, entity relationships, and structural constraints such as physical fault-propagation rules). The shared metadata records missing evidence, confounders, and constraint applicability rather than additional case labels.

Conditioned on $(x,\mu_x)$, the student policy $\pi_\theta$ produces a structured trajectory $\tau=(z_1,\ldots,z_H)$. Each transition $z_h$ belongs to $g(h)\in\{\textsc{Evidence},\textsc{Mechanism},\textsc{Conclusion}\}$, following $\textsc{Evidence}\prec\textsc{Mechanism}\prec\textsc{Conclusion}$ and $g(h)\preceq g(h+1)$. These abstract stages are shared across domains, while their entities, rules, relations, and vocabularies are domain-specific.

A retrieval function selects $\mathcal{K}_x\subseteq\mathcal{K}$ per case. Constraint applicability and evidence availability are evaluated jointly against $x$ and $\mu_x$. The teacher's coverage-aware evaluator, operating on this shared context together with $p_x$, judges every transition given its prefix:
\begin{equation}
V_{\mathcal{K}}(z_h \mid x, z_{<h}, \mu_x) \in \{\textit{verified}, \textit{violated}, \textit{unresolved}\}.
\label{eq:verifier}
\end{equation}
A transition is \textit{verified} when applicable constraints cover and support it, \textit{violated} when an applicable constraint is contradicted, and \textit{unresolved} when coverage or evidence is insufficient or applicable constraints conflict; the third state prevents such cases from becoming false supervision. The first wrong step is
\begin{equation}
h^{*} = \min\{\, h : V_{\mathcal{K}}(z_h \mid x, z_{<h}, \mu_x) = \textit{violated} \,\}.
\label{eq:fws}
\end{equation}
A trajectory enters localized correction only when
\begin{equation}
V_{\mathcal{K}}(z_j \mid x, z_{<j}, \mu_x) = \textit{verified}, \qquad \forall\, j<h^{*}.
\label{eq:eligible}
\end{equation}
The retained prefix $z_{<h^{*}}$ is verified only under the available applicable constraints. Trajectories containing unresolved prefix transitions are excluded from the main localized objective; a hybrid variant that routes them to auxiliary teacher judgment is evaluated separately.

For an eligible trajectory, localization constructs
\begin{equation}
s^{*} = \bigl((x,\mu_x),\; z_{<h^{*}},\; c^{*},\; \rho^{*},\; g(h^{*})\bigr),
\label{eq:state}
\end{equation}
where $c^{*}$ is the violated constraint and $\rho^{*}$ contains only the implicated entities, evidence, and relation needed for local repair. The teacher-side pipeline uses $y$ for evaluating cold-start candidates and validating repairs, but neither $y$ nor a complete teacher answer is exposed through $s^{*}$ or available to the student at inference time.

\paragraph{Structured transitions.}
Constraint-based verification requires inspectable causal transitions, whereas free-form chain-of-thought may mix evidence, inference, and conclusions in unconstrained prose. CausalOPD represents each transition as a typed tuple $z_h=(g_h, e_h, r_h, d_h)$: its causal stage, supporting observations, invoked rule or relation, and resulting decision. Every element required by a typed constraint thus occupies a declared field, so the evaluator can directly determine whether cited evidence supports the asserted relation and decision.
Missing fields, ambiguous references, or unavailable evidence produce \textit{unresolved} instead of a fabricated judgment.

\subsection{Knowledge-Augmented Teacher}

The knowledge-augmented teacher is a training-time supervision system with constructor, evaluator, and localizer interfaces: the evaluator and localizer combine the shared case context with $p_x$, with $\mathcal{K}_x$ retrieved into their context. During cold-start construction, privileged information reaches the constructor only through evaluator feedback. The design exploits an asymmetry: verifying steps against explicit criteria with $y$ available is far easier than the forward reasoning the student must learn.
The \emph{constructor} drafts expert trajectories from the same case context the student sees ($x$, $\mu_x$); revision toward $y$ is driven entirely by the evaluator's feedback below.
The \emph{evaluator} implements Eq.~\eqref{eq:verifier}, checking transition validity, evidence grounding, causal coherence, and conclusion consistency. The \emph{localizer} applies these judgments to student trajectories to identify $c^{*}$, $h^{*}$, and the corresponding repair context.

Trajectory construction and evaluation form an internal construct--evaluate--revise loop. Evaluator feedback flags unsupported evidence, invalid transitions, or inconsistent conclusions and guides targeted revision. A trajectory enters the cold-start set only when every transition is verified and its conclusion agrees with $y$; irreconcilable evidence--label conflicts are instead routed as data problems. This quality-control loop makes accepted trajectories credible references rather than merely fluent demonstrations.

The teacher proposes trajectories, whereas its evaluator determines their acceptance. Judgment is grounded in the explicit knowledge $\mathcal{K}_x$ rather than solely in the teacher's implicit parameters, and generation is disciplined by that judgment through the loop.
Accepted teacher trajectories provide repair evidence and a trusted reference, while constraint contradiction, rather than reference deviation, determines whether a student transition is violated. Cold-start SFT on accepted trajectories initializes knowledge-consistent generation and anchors subsequent on-policy optimization.

\subsection{First-Wrong-Step Correction}

After cold start, the student independently generates on-policy trajectories. The localizer evaluates each rollout in dependency order and returns the tri-state judgments, violated constraint identifier, and supporting evidence span. Equations~\eqref{eq:fws}--\eqref{eq:eligible} place the correction boundary between the verified prefix and affected suffix. This boundary distinguishes the originating violation from its propagated consequences without claiming that the prefix is globally error-free. A trajectory with no detected violation but an incorrect conclusion indicates insufficient effective verifier coverage and is excluded rather than treated as correct. Localization operates over causal transitions rather than tokens, matching the granularity at which domain constraints are stated and repair is meaningful.

Cold-start SFT teaches complete generation from $(x,\mu_x)$ but not resumption from the correction-state interface; correction-state SFT fills this gap. Given $s^{*}$ and $\mathcal{K}_x$, the constructor proposes repaired suffixes, while the evaluator uses $p_x$ to retain only candidates that repair $c^{*}$, contain no violated or unresolved transition, complete the required stages, and reach a reference-consistent conclusion. The verified prefix remains fixed as conditioning context, and the loss is applied only to the repaired transition and continuation.

After correction-state SFT, the student explores alternative suffixes from the same state under
\begin{equation}
R \;=\; R_{\text{rep}} \;+\; \lambda_1 R_{\text{nv}} \;+\; \lambda_2 R_{\text{out}} \;+\; \lambda_3 R_{\text{comp}},
\label{eq:reward}
\end{equation}
whose terms reward an evaluator-accepted repair at the boundary, continuation consistency, conclusion quality, and structural completion. The consistency term $R_{\text{nv}}$ scores each suffix transition $+1$ if verified, $-1$ if violated, and $0$ if unresolved, normalized over the suffix length; unresolved transitions thus receive no positive credit, preventing the policy from earning consistency reward outside verifier coverage.

A standard group-relative optimizer performs the update. The methodological design lies in translating a knowledge-verified boundary into focused exploration and credit assignment: the accepted prefix is preserved, the optimization horizon begins at $h^{*}$, and supervision concentrates on the suffix that actually requires correction rather than re-optimizing the entire response.

\subsection{Causal-Stage Curriculum}

Localization determines where each repair begins; the curriculum determines when repairs at each causal stage enter training. Correcting prerequisite-stage violations exposes eligible failures in later dependent stages, so a static mixture of correction states becomes increasingly misaligned with the student's learning needs. CausalOPD groups states by $g(h^{*})$ and advances the training focus from evidence identification to mechanism inference and conclusion attribution.

The focus advances when the incidence of first wrong steps at the active stage on a held-out probe set falls below a threshold, earlier-stage violations remain within tolerance, and the unresolved rate is bounded; the latter prevents reduced verification coverage from being counted as progress. Earlier-stage states from refreshed pools remain in training at a decaying rehearsal ratio to limit forgetting. Unlike difficulty- or competence-based curricula that rank training inputs along scalar axes, CausalOPD orders localized correction targets according to the dependency structure of the causal chain.

\paragraph{Iterative on-policy distillation.}
Unlike offline distillation from a fixed teacher corpus, CausalOPD reconstructs supervision under the current student policy. At round $t$, fresh trajectories sampled from $\pi_{\theta_t}$ are localized, the stage-specific correction pools are rebuilt, and localized suffix updates yield $\pi_{\theta_{t+1}}$; stale correction states are discarded. Supervision therefore tracks the student's evolving failure distribution rather than a fixed teacher-data distribution; the curriculum sets stage priorities each round.

\paragraph{Inference.}
At inference, the trained student conditions on $(x,\mu_x)$ under the structured output schema; the privileged teacher, reference conclusions, evaluator, localizer, correction-state construction, and curriculum scheduler are training-time components, and filtering rules, reward details, thresholds, and pseudocode appear in the technical appendix.

\section{Experiments}

\subsection{Experimental Setup}

\begin{table*}[t]
\centering
\footnotesize
\begin{tabularx}{\textwidth}{Xcccccccc}
\toprule
& \multicolumn{2}{c}{\textbf{Industrial}} & \multicolumn{2}{c}{\textbf{Clinical}} & \multicolumn{2}{c}{\textbf{Legal}} & \multicolumn{2}{c}{\textbf{Avg.}} \\
\cmidrule(lr){2-3}\cmidrule(lr){4-5}\cmidrule(lr){6-7}\cmidrule(lr){8-9}
Method & Path & Acc & Path & Acc & Path & Acc & Path & Acc \\
\midrule
\multicolumn{9}{c}{\textbf{Closed-Source LLMs (zero-shot)}} \\
\midrule
Qwen3.7-max (Teacher)      & 72.96 & 79.27 & 56.91 & 80.41 & 73.38 & 79.50 & 67.75 & 79.73 \\
Claude-Sonnet-4.6          & 74.68 & 87.54 & 59.91 & 84.09 & 79.50 & 90.29 & 71.36 & 87.31 \\
\midrule
\multicolumn{9}{c}{\textbf{Student: Qwen3-8B}} \\
\midrule
Qwen3-8B (zero-shot)       & 23.01 & 32.32 & 20.52 & 35.87 & 35.25 & 44.60 & 26.26 & 37.60 \\
Trajectory SFT & 62.61 & 76.26 & 49.40 & 68.17 & 45.68 & 60.07 & 52.56 & 68.17 \\
Outcome-only RL & 57.79 & 76.41 & 44.90 & 72.09 & 41.73 & 62.95 & 48.14 & 70.48 \\
Seq.-level OPD & 64.99 & 76.71 & 51.93 & 71.17 & 62.23 & 70.86 & 59.72 & 72.91 \\
Full-traj.\ process RL & 70.32 & 77.93 & 53.90 & 78.83 & 69.06 & 74.46 & 64.43 & 77.07 \\
\rowcolor{ourshade}
\textbf{CausalOPD}         & \textbf{93.78} & \textbf{96.07} & \textbf{74.04} & \textbf{81.57} & \textbf{81.65} & \textbf{88.13} & \textbf{83.16} & \textbf{88.59} \\
\bottomrule
\end{tabularx}
\caption{Main results (\%): strict path correctness (Path) and conclusion accuracy (Acc); Avg.: unweighted three-domain mean; best trained method in bold. Closed-source references: same schema, parser, and evaluation protocol (appendix).}
\label{tab:main}
\end{table*}

\paragraph{Benchmarks.}
We evaluate on three step-dependent causal reasoning benchmarks. \emph{Industrial}: air-handling-unit (AHU) fault diagnosis, where the model detects anomalous symptoms in multi-sensor operating records and infers the faulty component and root cause. Cases from the LBNL FDD dual-duct AHU dataset \citep{granderson2022lbnl} are split 80/20 into training (with an internal development subset) and held-out test cases. The industrial test set (3,942 cases) adds two cross-system sets never used for training or selection, the LBNL single-duct AHU and ASHRAE RP-1312 \citep{wen2011rp1312}, which differ in control strategy, sensor distribution, and fault assignment; cross-system claims rest on these subsets, reported separately (appendix), following the cross-system protocol of \citet{appliedenergy2024fault,aei2026fault}.
The conclusion space is open-vocabulary and includes fault labels unseen during task-specific training; scoring rules and seen/unseen breakdowns appear in the appendix.
\emph{Clinical}: diagnostic reasoning cases derived from the respiratory subset of DDXPlus \citep{tchango2022ddxplus}, following its official train/validation/test split (52,270 test cases).
\emph{Legal}: judgment reasoning cases derived from the MSLR insider-trading benchmark \citep{yu2025mslr}; lacking an official training split, its cases are partitioned 80/20 by case identity (278 test cases), stratified by year and violation type; all trained methods share the training split, inside which a development subset is held out for model and threshold selection.
All three domains share the evidence-identification, mechanism-inference, and conclusion-attribution stages with domain-specific knowledge bases and vocabularies. The settings stress complementary difficulty: cross-system transfer with partially unseen fault labels (industrial), an 18-pathology label space under a 246-fold class imbalance (clinical), and evidence-to-rule attribution over IRAC traces (legal); per-domain scope is tabulated in the appendix. To prevent leakage, splits are fixed by source case, system, and time window \emph{before} any derived artifact is generated, and knowledge sources and constraint templates are fixed without access to test conclusions.

\paragraph{Models and training.}
The student is Qwen3-8B; the teacher is the frozen Qwen3.7-max model, over an order of magnitude larger than the student, given the privileged context $p_x$.
Each domain trains an independent student under the same three-phase procedure with per-round fresh rollouts and correction-pool rebuilding; data, knowledge bases, and checkpoints are never shared across domains. All trained methods share the student base, prompts, output schema, and decoding configuration; hyperparameters, a 14B student-scale study, and a teacher-replacement study appear in the appendix.

\paragraph{Baselines.}
We compare CausalOPD with five controlled baselines: (i) the base student (zero-shot); (ii) trajectory SFT, i.e., off-policy imitation of teacher trajectories; (iii) outcome-only RL, which rewards final conclusions without process terms; (iv) API-compatible sequence-level OPD, which trains on teacher-corrected student rollouts under the same initialization, rollout count, update steps, and optimized-token budget, but without localization or stage-wise scheduling;
and (v) full-trajectory process RL, which optimizes the same reward terms as CausalOPD over complete responses. Closed-source LLMs are evaluated zero-shot under the identical protocol; an additional proprietary reference (GPT-4o) is reported in the appendix.

\paragraph{Metrics.}
\emph{Conclusion accuracy} checks the final attribution against $y$.
\emph{Path correctness} is strict: a trajectory counts only if every transition is judged verified, so violated and unresolved transitions both fail. Structurally incomplete outputs, including those missing a required stage or field, are treated as unresolved and also fail Path. Path scoring is anchored to per-case gold reasoning chains constructed separately from the training criteria: clinical chains are derived from benchmark-provided evidence and diagnosis annotations, legal chains from benchmark-provided IRAC traces, and industrial chains through expert adjudication; the adjudicating professionals repair and validate teacher-drafted chains against case data and benchmark documentation, without consulting the training knowledge base or constraint templates. Among complete outputs, semantic variants and constraint-consistent alternative paths are accepted; a path fails when it contains a confirmed contradiction. Gold chains are used only for evaluation and are never exposed to the training verifier.
The \emph{right-label--wrong-reasoning} (RLWR) rate is the fraction of correct conclusions whose trajectory contains at least one confirmed violation; a stricter variant counting unresolved transitions appears in the appendix. Training uses the criteria-driven verifier; test-time path evaluation uses a gold-chain-conditioned frozen scorer audited against method-blinded human judgments on test outputs (Sec.~4.3), and the verifier is itself validated against the gold chains below.

\paragraph{Human audit and statistics.}
Reliability is assessed by checking the training verifier's step judgments and first-wrong-step localizations against the gold chains on training-split rollouts, with domain professionals adjudicating semantic edge cases.
All reported accuracies average five sampled evaluation runs (temperature 0.6) per seed; we report means over three seeds with 95\% case-level bootstrap confidence intervals and McNemar's tests for paired binary outcomes.

\subsection{Main Results}

Table~\ref{tab:main} yields four findings. First, CausalOPD improves path correctness by 30.6 pp over trajectory SFT and 23.4 pp over sequence-level OPD averaged across domains; the gain persists on the two cross-system AHU subsets (34.6 pp over trajectory SFT), making system-specific memorization an insufficient explanation for the full gain.
Second, the RLWR rate falls from 19.1\% under trajectory SFT and 15.7\% under sequence-level OPD to 4.4\%: baselines frequently reach correct conclusions through invalid paths.
Third, full-trajectory process RL with the same reward terms recovers only 38.8\% of the path-correctness gain, indicating that localized short-horizon optimization, rather than reward design alone, drives the improvement.
Fourth, the students surpass both closed-source references on path correctness in all three domains and their zero-shot teacher on every metric; Claude-Sonnet-4.6 retains higher clinical and legal conclusion accuracy. The industrial accuracy levels are consistent with those reported for domain-fine-tuned HVAC diagnosis models \citep{appliedenergy2024fault,aei2026fault}.
Average Path reaches 83.16\% (95\% case-level bootstrap CI [82.74, 83.58]; legal [79.73, 83.57]); all nine per-domain Path comparisons against trajectory SFT, sequence-level OPD, and full-trajectory process RL remain significant after Holm correction (adjusted $p\leq 0.003$). Output-length and parsing controls are in the appendix.

\subsection{Supervision Reliability}

Table~\ref{tab:reliability} compares criteria-driven verification against independently constructed benchmark gold chains never used in training.

First, criteria$+y$ reaches 97.68--99.24\% agreement (85.8--91.0\% on the hardest wrong-conclusion test subset; appendix), above-95\% violation precision and recall, and 97.83--99.06\% coverage; the remainder are unresolved.

Second, first-wrong-step exact accuracy is 96.92--98.47\% against gold (stage-level accuracy and false-violation rate in appendix), anchoring short-horizon optimization at externally confirmed originating errors.

Third, supplying the same knowledge base at inference lifts the student only to 41.2\% average path correctness (KB-equipped teacher: 78.3\%) versus 83.2\% after distillation (appendix): reliable backward verification does not translate into forward competence without distillation.

On 150 held-out test trajectories per domain, three method-blinded raters agree with the automatic path judgments on 90.8--92.4\% and reproduce the method ordering (Kendall $\tau=1.0$; gaps likewise preserved under a cross-family scorer, appendix), supporting Table~\ref{tab:main}'s gains as reasoning improvements rather than scorer-family artifacts.

\begin{table}[!h]
\centering
\footnotesize
\begin{tabularx}{\columnwidth}{lXXX}
\toprule
Metric & Ind. & Clin. & Legal \\
\midrule
Violation recall                     & 97.80 & 95.90 & 96.90 \\
\rowcolor{ourshade}
Chain agreement, criteria$+y$        & \textbf{99.24} & \textbf{97.68} & \textbf{98.85} \\
Violation precision                  & 99.02 & 97.31 & 98.57 \\
FWS exact accuracy                   & 98.47 & 96.92 & 97.88 \\
Coverage                             & 99.06 & 97.83 & 98.92 \\
\bottomrule
\end{tabularx}
\caption{Reliability of criteria-driven supervision (\%) against gold chains over 38,712/69,388/6,467 transitions (Ind./Clin./Legal). FWS accuracy: trajectories with a gold first wrong step; coverage: resolved (non-abstained) fraction. Class-balanced, hard-subset, and knowledge-ablation metrics in appendix.}
\label{tab:reliability}
\end{table}

\subsection{Ablation Study}

\begin{table}[!b]
\centering
\footnotesize
\begin{tabularx}{\columnwidth}{Xcccc}
\toprule
Variant & Path & Acc & RLWR$\downarrow$ & Tok.$\downarrow$ \\
\midrule
\rowcolor{ourshade}
\textbf{CausalOPD (full)}   & \textbf{83.16} & \textbf{88.59} & \textbf{4.40} & 0.63 \\
w/o knowl.\ grounding       & 60.69 & 73.90 & 15.83 & 0.71 \\
w/o teacher revision        & 54.24 & 69.28 & 18.41 & 0.73 \\
w/o FWS (full response)     & 64.43 & 77.07 & 9.22  & 1.00 \\
w/o correction-state SFT    & 77.26 & 84.99 & 6.81  & 0.65 \\
w/o curriculum (joint)      & 79.21 & 86.12 & 5.74  & 0.64 \\
w/o fresh refresh (stale)   & 74.18 & 82.69 & 6.99  & 0.67 \\
\bottomrule
\end{tabularx}
\caption{Ablations (\%, unweighted three-domain mean). Tok.: completion tokens optimized per update, normalized to the full-response variant ($=1.00$); prompt tokens excluded. Variant Tok.\ values are weighted by each variant's FWS distribution (appendix); w/o FWS coincides with full-trajectory process RL in Table~\ref{tab:main}.}
\label{tab:ablation}
\end{table}

Table~\ref{tab:ablation} isolates each mechanism. Replacing knowledge-grounded evaluation with the teacher's implicit judgment costs 22.5 pp of path correctness and raises RLWR by 11.4 pp, collapsing performance toward the sequence-level OPD tier; cold-starting from unrevised teacher trajectories (no construct--evaluate--revise) costs 28.9 pp, close to the trajectory-SFT tier. Teacher revision is the largest single ablation, showing that supervision quality is foundational; among the online correction components, FWS localization has the largest effect. Both the revision and stale-pool ablations match cold-start volume, training tokens, update steps, optimizer, and curriculum, isolating quality from quantity.

Optimizing the full response from the same reward terms instead of the localized suffix forfeits 18.7 pp while consuming 1.6 times more optimized tokens, and skipping correction-state SFT costs a further 5.9 pp: the first-wrong-step boundary converts process rewards into targeted repair.
Random-step localization recovers only 41\% of the FWS gain (appendix): boundary placement matters beyond merely shortening the optimization horizon.

Joint mixing of all stages forfeits 4.0 pp relative to the causal-stage curriculum, and freezing the round-1 pool under a matched budget forfeits 9.0 pp: the gain comes from rebuilding supervision under the current policy. Replacing the causal stage order with the reverse order under the same budget forfeits 16.4 pp, falling below even joint mixing (appendix), so the gain traces to the causal ordering itself rather than stage-wise scheduling alone. A weaker-teacher replacement preserves the gain (macro Path 82.46, 22.7 pp over sequence-level OPD; appendix).

\subsection{Iterative Behavior and Efficiency}

Figure~\ref{fig:analysis} reports industrial dynamics (other domains in appendix). Fresh-rollout OPD rises from 65.0\% to 98.0\% over five rounds; the budget-matched stale pool plateaus after round 3 and ends 9.3 pp lower. Evidence-stage and total first-wrong-step incidence fall from 21.0\% to 0.3\% and from 35.0\% to 2.0\%, while conclusion-stage share among remaining errors rises from 11\% to 33\%. CausalOPD optimizes 3.61M completion tokens (62.6\% of full-trajectory RL's 5.76M); prefix retention cuts the mechanism-stage share from 75.3\% to 41.9\%. Costs exclude rollout generation, verifier/teacher calls, prompts, and wall time (appendix).

\par\smallskip
\noindent\begin{minipage}{\columnwidth}
\centering
\includegraphics[width=\columnwidth]{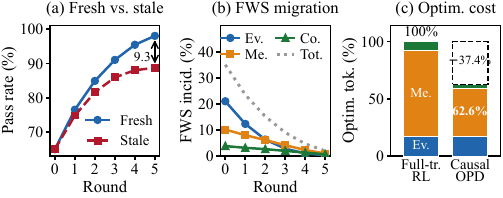}
\captionof{figure}{Iterative behavior (industrial). (a) Fresh vs.\ budget-matched stale pools on a held-out 3,569-case progress set. (b) Per-stage FWS incidence on a 600-trajectory probe subset. (c) Optimized completion tokens by stage. Both development sets are test-disjoint; the probe controls curriculum transitions.}
\label{fig:analysis}
\end{minipage}

\section{Conclusion}

We presented CausalOPD, combining knowledge-grounded teacher revision, tri-state first-wrong-step localization, and curriculum-guided suffix RL on fresh rollouts. Across three domains, 8B students improve paths and conclusions over trained baselines, reduce the right-label--wrong-reasoning (RLWR) rate, surpass proprietary references on paths, and retain cross-system industrial gains. These results support deploying the 8B students locally on the studied tasks, but verification certifies only curated knowledge. Future work includes reducing knowledge-curation cost, operating under weaker knowledge bases, and extending the schema to less structured domains.

\bibliography{references}

\end{document}